%% file: ms.tex
\documentclass[12pt]{article}

\usepackage{geometry}
\usepackage[utf8]{inputenc}
\usepackage[T1]{fontenc}
\usepackage{url} 
\usepackage{booktabs} 
\usepackage{amsfonts} 
\usepackage{nicefrac}  
\usepackage{microtype}  
\usepackage{graphicx}
\usepackage{xcolor}

\usepackage{amsthm}
\usepackage{amsmath}
\usepackage{amssymb}
\usepackage{multirow}
\usepackage{bbm}
\usepackage{enumitem}
\usepackage{wrapfig}
\usepackage{epsfig,psfrag}
\usepackage{caption}
\usepackage{subcaption}
\usepackage[normalem]{ulem}
\usepackage{bm}
\usepackage{authblk}
\usepackage{natbib}
\usepackage{changepage}
\usepackage{bbold}

\usepackage{algorithm}
\usepackage{algpseudocode}

\usepackage{hyperref}  
\hypersetup{colorlinks,
citecolor=blue,
filecolor=blue,
linkcolor=red,
urlcolor=blue}
\usepackage[capitalize,noabbrev]{cleveref}

\input{math_commands.tex}

\title{
\centering \large \textbf{Topological Continual Learning \\with Wasserstein Distance and Barycenter}
}

\author[1]{\normalsize Tananun Songdechakraiwut}
\author[2]{\normalsize Xiaoshuang Yin}
\author[1]{\normalsize Barry D. Van Veen}
\date{}
\affil[1]{\small University of Wisconsin--Madison}
\affil[2]{\small Google}

\begin{document}

\maketitle

\begin{abstract}
Continual learning in neural networks suffers from a phenomenon called catastrophic forgetting, in which a network quickly forgets what was learned in a previous task. The human brain, however, is able to continually learn new tasks and accumulate knowledge throughout life. Neuroscience findings suggest that continual learning success in the human brain is potentially associated with its modular structure and memory consolidation mechanisms. In this paper we propose a novel topological regularization that penalizes cycle structure in a neural network during training using principled theory from persistent homology and optimal transport. The penalty encourages the network to learn modular structure during training. The penalization is based on the closed-form expressions of the Wasserstein distance and barycenter for the topological features of a 1-skeleton representation for the network. Our topological continual learning method combines the proposed regularization with a tiny episodic memory to mitigate forgetting. We demonstrate that our method is effective in both shallow and deep network architectures for multiple image classification datasets.
\end{abstract}

\newpage

\section{Introduction}

Neural networks can be trained to achieve impressive performance on a variety of learning tasks. However, when an already trained network is further trained on a new task, a phenomenon called \emph{catastrophic forgetting} \citep{mccloskey1989catastrophic} occurs, in which previously learned tasks are quickly forgotten with additional training.
The human brain, however, is able to continually learn new tasks and accumulate knowledge throughout life without significant loss of previously learned skills.
The biological mechanisms behind this trait are not yet fully understood.
Neuroscience findings suggest that the principle of modularity \citep{hart2010neural} may play an important role. Modular structures \citep{simon1962architecture} are aggregates of modules (components) that perform specific functions without perturbing one another. Human brains have been characterized by modular structures during learning \citep{finc2020dynamic}. Such structures are hypothesized to reduce the interdependence of components, enhance robustness, and facilitate learning \citep{bassett2011dynamic}.
Another important learning mechanism is the hippocampal and neocortical interaction responsible for memory consolidation \citep{mcgaugh2000memory}. In particular, the hippocampal system encodes recent experiences that later are replayed multiple times before consolidation as episodic memory in the neocortex \citep{klinzing2019mechanisms}. The interplay between the modularity principle and memory consolidation, among other mechanisms, is potentially associated with continual learning success.

Persistent homology \citep{barannikov1994framed,edelsbrunner2000topological,wasserman2018topological} has emerged as a tool for understanding, characterizing and quantifying the topology of brain networks.
For example, it has been used to evaluate biomarkers of the neural basis of consciousness \citep{songdechakraiwut2022fast} and the impact of twin genetics \citep{songdechakraiwut2021topological} by interpreting brain networks as 1-skeletons \citep{munkres1996elements} of a simplicial complex.
The topology of a 1-skeleton is \emph{completely} characterized by connected components and cycles \citep{songdechakraiwut2022fast}.
Brain networks naturally organize into modules or connected components \citep{bullmore2009complex,honey2007network}, while cycle structure is ubiquitous and is often interpreted in terms of information propagation, redundancy and feedback loops \citep{kwon2007analysis,ozbudak2005system,venkatesh2004multiple,weiner2002ptdinsp}.

\begin{figure}[t]
\centering
\centerline{\includegraphics[width=.5\columnwidth]{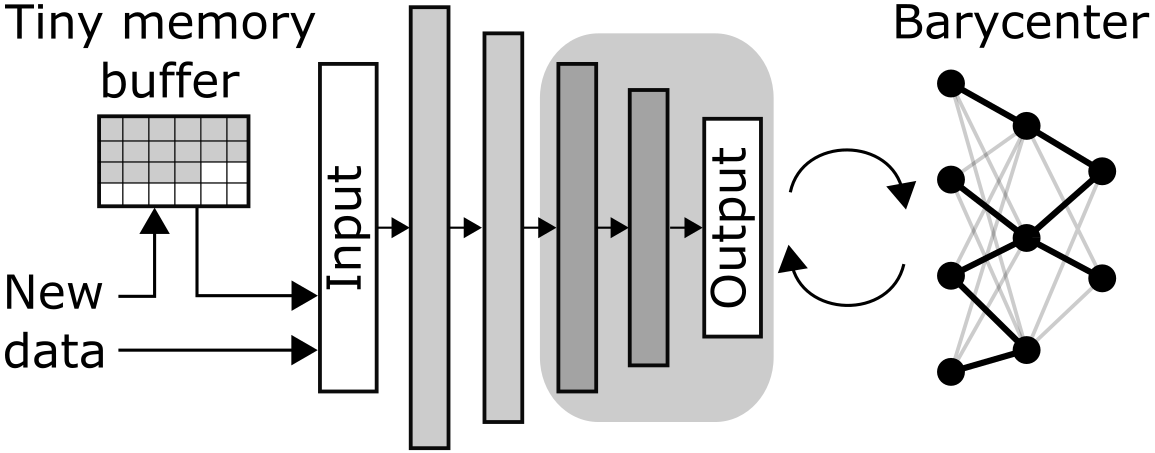}}
\caption{Schematic illustrating our topological continual learning approach. A tiny episodic memory replays past examples from previously learned tasks. The cycle topology of a subset of layers (shaded) is regularized based on the Wasserstein distance between the network and barycenter of previously learned networks to improve the generalization of the learning and memory consolidation processes.}
\label{fig:schematic}
\end{figure}

In this paper we show that persistent homology can be used to improve performance of neural networks in continual learning tasks. In particular, we interpret a network as a 1-skeleton and propose a novel \emph{topological regularization} of the 1-skeleton's cycle structure to avoid catastrophic forgetting of previously learned tasks.
Regularizing the cycle structure allows the network to explicitly learn its complement, i.e., the modular structure, through gradient optimization.
Our approach is made computationally efficient by use of the closed form expressions for the Wasserstein barycenter and the gradient of Wasserstein distance between network cycle structures presented in \citep{songdechakraiwut2021topological,songdechakraiwut2022fast}. 
\cref{fig:schematic} illustrates that our proposed approach employs topological regularization with a tiny episodic memory to mitigate forgetting and facilitate learning. We evaluate our approach using image classification across multiple data sets and show that it generally improves classification performance compared to competing approaches in the challenging case of both shallow and deep networks of limited width when faced with many learning tasks.

The paper is organized as follows. Efficient computation of persistent homology for neural networks is given in \cref{sec:comptop}, and \cref{sec:method} presents our topology-regularized continual learning strategy. 
In \cref{sec:experiment}, we compare our methods to multiple baseline algorithms using four image classification datasets. \cref{sec:related} provides a brief discussion of the relationship between the proposed approach and previously reported methods for continual learning.

\section{Efficient Computation of Topology for Network Graphs}
\label{sec:comptop}

\subsection{Graph Filtration}

\begin{figure}[t]
\centering
\centerline{\includegraphics[width=\columnwidth]{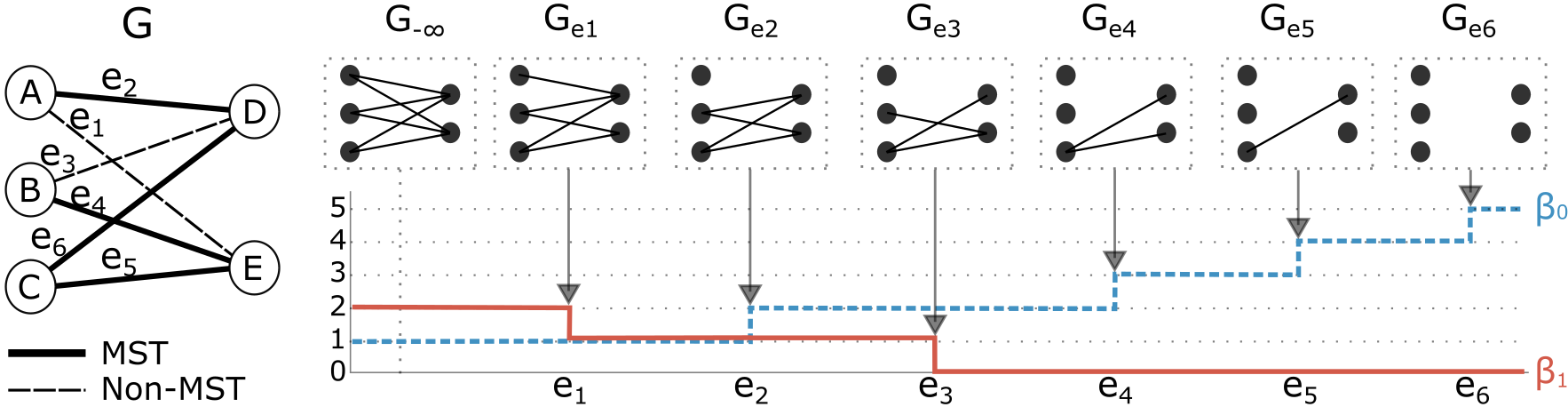}}
\caption{Illustration of graph filtration. A toy neural network $G$ showing its maximum spanning tree (MST) as edges denoted by dark lines and a subnetwork with non-MST edges shown as dashed lines. As the filtration value increases, the number of connected components $\beta_0$ monotonically increases while the number of cycles $\beta_1$ monotonically decreases. Connected components are born at the MST edge weights $e_2,e_4,e_5,e_6$ while cycles die at the non-MST edge weights $e_1,e_3$.}
\label{fig:ph}
\end{figure}

Represent a neural network as an undirected weighted graph $G=(V,\sW)$ with a set of nodes $V$, and a set of edge weights $\sW=\{w_{i,j}\}$. The number of nodes and weights are denoted by $|V|$ and $|\sW|$, respectively.
Create a binary graph $G_\epsilon$ with the identical node set $V$ by thresholding the edge weights so that an edge  between nodes $i$ and $j$ exists if $w_{i,j} > \epsilon$. The binary graph is a simplicial complex consisting of only nodes and edges known as a \emph{1-skeleton} \citep{munkres1996elements}.
As $\epsilon$ increases, more and more edges are removed from the network $G$, resulting in a nested sequence of 1-skeletons:
\begin{equation}
\label{eq:graphfiltration}
G_{\epsilon_0} \supseteq G_{\epsilon_1} \supseteq \cdots \supseteq G_{\epsilon_k} ,
\end{equation}
where $\epsilon_0 \leq \epsilon_1 \leq \cdots \leq \epsilon_k$ are called filtration values. This sequence of 1-skeletons is called a \emph{graph filtration} \citep{lee2012persistent}.
\cref{fig:ph} illustrates the graph filtration of a toy neural network.

\subsection{Birth and Death Decomposition}
\label{sec:decomp}

The only non-trivial topological features in a 1-skeleton are \emph{connected components} (0-dimensional topological features) and \emph{cycles} (1-dimensional topological features). There are no higher-dimensional topological features in the 1-skeleton, in contrast to clique complexes \citep{otter2017roadmap} and Rips complexes \citep{ghrist2008barcodes,zomorodian2010fast}.
Persistent homology keeps track of the birth and death of topological features over filtration values $\epsilon$. If a topological feature is born at a filtration value $b_l$ and persists up to a filtration value $d_l$, then this feature is represented as a two-dimensional \emph{persistence} point $(b_l,d_l)$ in a plane. The set of all points $\{(b_l,d_l)\}_l$ is called \emph{persistence diagram} \citep{edelsbrunner2008persistent} or, equivalently, \emph{persistence barcode} \citep{ghrist2008barcodes}.
The use of the 1-skeleton simplifies the persistence barcodes to one-dimensional descriptors \citep{songdechakraiwut2021topological,songdechakraiwut2022fast}.
Specifically, the representation of the connected components can be simplified to a collection of \emph{birth values} $\sB(G) = \{b_l \}$ and that of cycles to a collection of \emph{death values} $\sD(G)=\{ d_l \}$. In addition, neural networks of the same architecture have a birth set $\sB$ and a death set $\sD$ of the \emph{same} cardinality as $|V| - 1$ and $|\sW| - (|V|-1)$, respectively.
This result completely resolves the problem of point mismatch in persistence barcodes for same-architecture neural networks.
The example network of Fig. \ref{fig:ph} has $\sW=\{e_i\}_{i=1}^6, \sB(G)=\{e_2,e_4,e_5,e_6\}$ and $\sD(G)=\{e_1,e_3\}$.
$\sB(G)$ and $\sD(G)$ can be identified very efficiently by computing the \emph{maximum spanning tree} (MST) of the network \citep{lee2012persistent} in $O(n\log n)$ operations, where $n$ is the number of edges in the network.
Supplementary description of the decomposition is provided in \cref{supp:decomp}.

\subsection{Closed Form Wasserstein Distance and Gradient}
The birth set of connected components and the death set of cycles are theoretically shown to completely characterize the topology of a 1-skeleton network representation as described in \cref{sec:decomp}.
The Wasserstein distance between 1-skeleton network representations has a closed-form expression \citep{songdechakraiwut2022fast}.
Here we only consider the Wasserstein distance for cycle structure, which depends solely on the death sets.
Let $G, H$ be two given networks based on the same architecture.
Their (squared) \emph{2-Wasserstein distance for cycles} is defined as the optimal matching cost between $\sD(G)$ and $\sD(H)$:
\begin{equation}
\label{eq:bdwass}
 W^2_{cycle}(G,H) = \min_{\phi} \sum_{d_l \in \sD(G)} \big[ d_l - \phi(d_l)\big]^2 ,
\end{equation}
where $\phi$ is a bijection from $\sD(G)$ to $\sD(H)$.
The Wasserstein distance form given in (\ref{eq:bdwass}) has a closed-form expression that allows for very efficient computation as follows \citep{songdechakraiwut2022fast}

\begin{equation}
\label{eq:1dwass}
W^2_{cycle}(G,H) = \sum_{d_l \in \sD(G)} \big[ d_l - \phi^*(d_l)\big]^2 ,
\end{equation}

where $\phi^*$ maps the $l$-th smallest death value in $\sD(G)$ to the $l$-th smallest death value in $\sD(H)$ for all $l$.

In addition, the gradient of the Wasserstein distance for cycles $\nabla_G W^2_{cycle}(G,H)$ with respect to edge weights $w_{i,j} \in \sW$ is given as a gradient matrix whose $i,j$-th entry is \citep{songdechakraiwut2022fast}
\begin{equation}
\label{eq:1dgrad}
\frac{\partial W^2_{cycle}(G,H)}{\partial w_{i,j}} 
= 
\begin{cases}
0 & \text{if } w_{i,j} \in \sB(G); \\
2 \big[ w_{i,j} - \phi^*( w_{i,j} ) \big] & \text{if } w_{i,j} \in \sD(G).
\end{cases}
\end{equation}
This follows because the edge weight set decomposes into the collection of births and the collection of deaths as described in \cref{sec:decomp}. Intuitively, by slightly adjusting the edge weight $w_{i,j}$ corresponding to a death value in $\sD(G)$, we in turn alter the cycle structure of the network $G$. The gradient form given in (\ref{eq:1dgrad}) can be equivalently expressed as
\begin{equation}
\label{eq:grad}
\frac{\partial W^2_{cycle}(G,H)}{\partial d_l} 
= 
2 \big[ d_l - \phi^*( d_l ) \big],
\end{equation}
where $d_l \in \sD(G)$ and $\phi^*$ is the optimal matching from $\sD(G)$ to $\sD(H)$.
The gradient computation requires sorting to find the optimal matching $\phi^*$, which requires $O(n\log n)$ operations, where $n$ is the number of edges in networks.

\subsection{Closed Form Wasserstein Barycenter}

The Wasserstein barycenter is the mean of a collection of networks under the Wasserstein distance and represents the topological centroid. Consider same-architecture networks $G^{(1)},...,G^{(N)}$. Their death sets have identical cardinality, i.e., $|\sD| = |\sD(G^{(1)})| = |\sD(G^{(2)})| = \cdots = |\sD(G^{(N)})|$ because the cardinality depends only on the number of nodes and edge weights. The \emph{Wasserstein barycenter for cycles} $\mathcal{G}_{cycle}$ is defined as the death set that minimizes the \emph{weighted} sum of the Wasserstein distances for cycles:
\begin{equation}
\label{eq:topmeanopt}
\mathcal{G}_{cycle} = \argmin_\mathcal{G} \sum_{i=1}^N \nu_i W^2_{cycle}(\mathcal{G}, G^{(i)}) ,
\end{equation}
where $\nu_i$ is a non-negative weight corresponding to network $G^{(i)}$ that is used to emphasize the contribution of some networks to the final barycenter more than others.
The closed-form Wasserstein distance given in (\ref{eq:1dwass}) results in a closed form expression for the Wasserstein barycenter as follows.

Let $\sD(G^{(i)}): d^{(i)}_{1} \leq \cdots \leq d^{(i)}_{|\sD|}$ be the death set of network $G^{(i)}$.
It follows that the $l$-th smallest death value of the barycenter $\mathcal{G}_{cycle}$ of the $N$ networks is given by the weighted mean of all the $l$-th smallest death values of such networks, i.e., $\mathcal{G}_{cycle}: \overline d_1 \leq \cdots \leq \overline d_{|\sD|},$ where
\begin{equation}
\label{eq:topmeanclosedform}
\overline d_l = \sum_{i=1}^N \nu_i d^{(i)}_{l} \Big/ \sum_{i=1}^N \nu_i.
\end{equation}
\cref{eq:topmeanclosedform} is derived by setting the derivative of quadratic function in Eq. \ref{eq:topmeanopt} equal to zero.
The complete proof is given in \cref{supp:proof}.

\section{Topology Based Continual Learning}
\label{sec:method}

Consider a \emph{continual learning} scenario in which $T$ supervised learning tasks are learned sequentially. Each task has a task descriptor $\tau \in \sT = \{1,2,...,T\}$ with a corresponding dataset $\sP_\tau = \{(\vx_{i,\tau},\vy_{i,\tau})_{i=1}^{N_\tau}\}$ containing $N_\tau$ labeled training examples consisting of a feature vector $\vx_{i,\tau} \in \gX$ and a target vector $\vy_{i,\tau} \in \gY$.
We further consider the continuum of training examples that are experienced \emph{only once}, and assume that the continuum is \emph{locally independent and identically distributed} (iid), i.e., $(\vx_{i,\tau},\vy_{i,\tau}) \overset{iid}{\sim} P_{\tau}$ following the prior work of \citet{lopez2017gradient}. 
The goal is to train a model $f: \gX \times \sT \rightarrow \gY$ that predicts a target vector $\vy$ corresponding to a test pair $(\vx, \tau)$, where $(\vx, \vy) \sim P_\tau$.

Consider a neural network predictor $f$ parameterized by a weight set $\sW$. For the first task $\tau=1$, we find optimal weight set $\sW^*$ by minimizing the expectation of a loss function as:
\begin{equation}
\label{eq:expectloss}
    \sW^* = \argmin_\sW \E_{(\vx,\vy)\sim P_1} L\big(f(\sW; \vx, \tau=1), \vy \big),
\end{equation}
which is typically accomplished by employing the empirical risk minimization (ERM) principle \citep{vapnik1991principles}. That is, the minimization problem given in (\ref{eq:expectloss}) is approximated by minimizing the empirical loss function
\begin{equation}
    \gL_{ERM}(\sW) = \frac{1}{N_1} \sum_{i=1}^{N_1} L\big(f(\sW; \vx_{i,1}, \tau=1), \vy_{i,1} \big).
\end{equation}
This minimization can be effectively solved by gradient descent. However, extending this approach to more tasks causes catastrophic forgetting \citep{mccloskey1989catastrophic} of previously learned tasks. That is, the model forgets how to predict past tasks after it is exposed to future tasks. Our approach to addressing this problem is to topologically penalize training with future tasks based on the underlying 1-skeleton of the neural network.

Define a neural network $G^{(\tau)}$ for learning task $\tau$ with nodes given by neurons, and edge weights defined by the weight/parameter set $\sW$. 
All past-task networks $G^{(j)}$, for $j=1,...,\tau-1$, have the identical node sets with the trained weight set $\sW^*_j$ denoting the weights after training through the entire sequence up to task $j$. 
Since these graphs have the same architecture, their death sets have the same cardinality denoted by $|\sD|$.
Then the birth-death decomposition of the weight set $\sW^*_j$ results in the death set $\sD(G^{(j)}): d^{(j)}_{1} \leq \cdots \leq d^{(j)}_{|\sD|}$.
The Wasserstein barycenter for cycles of the first $\tau-1$ training tasks associated with networks $G^{(1)},...,G^{(\tau-1)}$ is
\begin{equation}
\label{eq:barycenter}
    \gG_{cycle}^{(\tau-1)}: \overline d^{(\tau-1)}_{1} \leq \cdots \leq \overline d^{(\tau-1)}_{|\sD|},
\end{equation}
where $\overline d^{(\tau-1)}_{l} = \sum_{j=1}^{\tau-1} \nu_j d^{(j)}_{l} \Big/ \sum_{j=1}^{\tau-1} \nu_j$. Our approach to learning task $\tau$ minimizes the ERM loss with the Wasserstein distance and barycenter penalty
\begin{equation}
\label{eq:clloss}
    \gL_\tau(\sW) = \gL_{ERM,\tau}(\sW) + \frac{\lambda}{2} W^2_{cycle}(G^{(\tau)},\gG_{cycle}^{(\tau-1)}) \quad \text{for all task } \tau>1,
\end{equation}
where $\lambda$ controls relative importance between past- and current-task cycle structure. 
Intuitively, we penalize changes of cycle structure in a neural network while allow the network to explicitly learn the modular structure. 

Minimization of Eq. (\ref{eq:clloss}) is accomplished via gradient descent over all training samples in learning task $\tau$. 
Direct application of the gradient form given in (\ref{eq:grad}) results in
\begin{equation}
\label{eq:cl1dgrad}
\frac{\partial W^2_{cycle}(G^{(\tau)},\gG_{cycle}^{(\tau-1)})}{\partial d^{(\tau)}_{l}}
= 2 \big[ d^{(\tau)}_{l} - \phi^*( d^{(\tau)}_{l} ) \big],
\end{equation}
where $d^{(\tau)}_{l} \in \sD(G^{(\tau)})$ and $\phi^*$ is the optimal matching from $\sD(G^{(\tau)})$ to $\gG_{cycle}^{(\tau-1)}$. In other words, the gradient computation is achieved in two steps: 1) compute birth-death decomposition of $\sW$ to find $\sD(G^{(\tau)})$; 2) sort $\sD(G^{(\tau)})$ to find the optimal matching $\phi^*$ between $\sD(G^{(\tau)})$ and $\gG_{cycle}^{(\tau-1)}$. 
Furthermore, we execute the first step every $m$ iterations. 
When $m=1$, we compute birth-death decomposition every iteration to determine which edge belongs to $\sD(G^{(\tau)})$ before sorting, while $m>1$ allows to utilize multiple gradient descent updates to consolidate changes to previously determined edges in $\sD(G^{(\tau)})$ by directly sorting updated weights in $\sD(G^{(\tau)})$ from previous iterations. This approach may mimic human learning and memory consolidation over multiple temporal scales from days to decades \citep{bassett2011dynamic}. Thus, the computational complexity for gradient computation is $n \log n$ operations, where $n$ is the number of weights/parameters in the neural network graph.

At any time, we only need to store \emph{one} Wasserstein barycenter from the previous task $\gG_{cycle}^{(\tau-1)} = \{\overline d^{(\tau-1)}_{l}\}_l$. The barycenter for the current task $\gG_{cycle}^{(\tau)} = \{\overline d^{(\tau)}_{l}\}_l$ can be computed as:
\begin{equation}
\label{eq:topmeanupdate}
    \overline d^{(\tau)}_{l} = \frac{p \overline d^{(\tau-1)}_{l} + q d^{(\tau)}_{l}}{p+q},
\end{equation}
where $d^{(\tau)}_{l} \in \sD(G^{(\tau)}(\sW^*_\tau))$ and $p,q>0$ used to emphasize the contribution of $\gG_{cycle}^{(\tau-1)}$ and $\sD(G^{(\tau)})$ to $\gG_{cycle}^{(\tau)}$. This online update satisfies the closed-form Wasserstein barycenter given in (\ref{eq:topmeanclosedform}). 
A proof by induction is provided in \cref{supp:proof}.

We summarize our topological continual learning steps in \cref{alg:topcl}.

\paragraph{Subgraph regularization}
We note that it is straightforward and may be advantageous to limit the topological regularization to specific layers and depths in the neural network. For example, a convolutional neural network is typically constructed as a series of convolutional (conv) layers followed by a fully-connected (fc) layer. The conv layers extract high-level features from data for the fc layer, which in turn uses those features for classification. We can define subgraphs based on groupings of a conv layer with the following fc layer and restrict the topological regularizer to a subset of layers, typically layers towards the output side of the neural network.
Formally, if we have $K$ subnetwork graphs $G^{(\tau,1)},...,G^{(\tau,K)}$ and their barycenters $\gG_{cycle}^{(\tau-1,1)},...,\gG_{cycle}^{(\tau-1,K)}$, then our topological penalty is expressed as the sum of Wasserstein distances for each separate graph, i.e.,
\begin{equation}
    \gL_\tau(\sW) = \gL_{ERM,\tau}(\sW) + \frac{\lambda}{2} \sum_{k=1}^K W^2_{cycle}(G^{(\tau,k)},\gG_{cycle}^{(\tau-1,k)}) \quad \text{for all task } \tau>1.
\end{equation}

\begin{algorithm}
\caption{Topological continual learning algorithm}
\label{alg:topcl}
\small
\begin{algorithmic}[1]
\Procedure{TOP}{$\sP, \sW=\{w\},\lambda,p,q,\gamma$}
\State $\sM \gets \{\}$ \Comment{Allocate a tiny memory buffer}
\For{$B_{\sP} \sim \sP_1$} \Comment{Sample without replacement a mini-batch from $1^{st}$ dataset}
    \State $g_{sgd} \gets \nabla_w \gL_{ERM}(B_\sP)$ \Comment{Compute gradient using the mini-batch}
    \State $w \gets w - \gamma \cdot g_{sgd}$
    \State $\sM \gets$ mem\_update$(\sM,B_{\sP})$ \Comment{Update memory; see \cref{sec:experiment}}
\EndFor
\State $\gG^{(1)} \gets \sD(\sW)$ \Comment{Initial barycenter described in \cref{eq:barycenter}}
\For{$\tau \in \{2,..,T\}$}
    \State $iter \gets 0$
    \For{$B_{\sP} \sim \sP_\tau$} \Comment{Sample without replacement a mini-batch from $\tau^{th}$ dataset}
        \State $B_\sM \sim \sM$ \Comment{Sample a mini-batch from the buffer}
        \If{$iter$ mod $m == 0$} \Comment{Every $m$ iterations}
            \State $\sD \gets$ bd\_decomposition$(\sW)$ \Comment{Compute death set from current weight set}
        \EndIf
        \State $g_{sgd} \gets \nabla_w \gL_{ERM}(B_\sP \cup B_\sM)$ \Comment{Compute gradient using aggregated mini-batch}
        \If{$w \in \sD$}
            \State $g_{top} \gets w - \phi^*(w)$ \Comment{Compute closed form gradient in \cref{eq:cl1dgrad}}
            \State $w \gets w - \gamma \cdot (g_{sgd} + \lambda \cdot g_{top})$
        \Else
            \State $w \gets w - \gamma \cdot g_{sgd}$
        \EndIf
        \State $iter \gets iter+1$
        \State $\sM \gets$ mem\_update$(\sM,B_{\sP})$
    \EndFor
    \State $\gG^{(\tau)} \gets \gG^{(\tau-1)}(p,q)$ \Comment{Update barycenter using \cref{eq:topmeanupdate}}
\EndFor
\State \Return $\sW$ \Comment{Return optimal weight set}
\EndProcedure
\end{algorithmic}
\end{algorithm}

\section{Image Classification Experiments}
\label{sec:experiment}

\paragraph{Datasets}
We perform continual learning experiments on four datasets.
(1) \emph{Permuted MNIST} (P-MNIST) \citep{kirkpatrick2017overcoming} is a variant of the MNIST dataset of handwriten digits \citep{lecun1998gradient} where each task is constructed by a fixed random permutation of image pixels in each original MNIST example.
(2) \emph{Rotated MNIST} (R-MNIST) \citep{lopez2017gradient} is another variant of the MNIST dataset where each task contains digits rotated by a fixed angle between 0 and 180 degrees.
Both P-MNIST and R-MNIST contain 30 sequential tasks, each has 10,000 training examples and 10 classes. Each task in P-MNIST and R-MNIST is constructed with different permutations and rotations, respectively, resulting in different distributions among tasks.
The other two datasets, (3) \emph{Split CIFAR} and (4) \emph{split miniImageNet}, are constructed by randomly splitting 100 classes in the original CIFAR-100 dataset \citep{krizhevsky2009learning} and ImageNet dataset \citep{russakovsky2015imagenet,vinyals2016matching} into 20 learning tasks each with 5 classes.

\paragraph{Network architecture}
We are interested in a challenging scenario of limited network width and a long sequence of learning tasks, i.e., the practical scenario in which an assumption of over-parameterized neural networks is relaxed.
Specifically, the P-MNIST and R-MNIST datasets use a fully-connected neural network with two hidden layers each with 128 neurons while the split CIFAR and split miniImageNet datasets use a downsized version of ResNet18 \citep{he2016deep} with eight times fewer feature maps across all layers, similar to \citep{lopez2017gradient}.
Note that the works in \citep{chaudhry2020continual,liu2022continual} require much larger neural networks to satisfy the over-parameterization assumption, which in turn require a whole lot more computational power and memory footprint to optimize their parameter sets to perform well.
For P-MNIST and R-MNIST, we evaluate classification performance in a \emph{single-head} setting where all tasks share the final classifier layer in the fully-connected network, and thus the inference is performed without task identifiers. For the split CIFAR and split miniImageNet datasets, the classification performance is evaluated in a \emph{multi-head} setting where each task has a separate classifier in the reduced ResNet18.

\paragraph{Method comparison}
We evaluate our method performance in relative to eight baseline approaches that learn a sequence of tasks in a fixed-size network architecture. (1) \emph{Finetune} is a model trained sequentially without any regularization and past-task episodic memory. (2) \emph{Elastic weight consolidation} (EWC) \citep{kirkpatrick2017overcoming} is a regularization-based method that penalizes changes in parameters that were important for the previous tasks using the Fisher information. (3) \emph{Recursive gradient optimization} (RGO) \citep{liu2022continual} combines use of gradient direction modification and task-specific random rearrangement to network layers to mitigate forgetting between tasks. (4) \emph{Averaged gradient episodic memory} (A-GEM) \citep{chaudhry2018efficient} and (5) ORTHOG-SUB \citep{chaudhry2020continual} methods store past-task examples in an episodic memory buffer used to project gradient updates to avoid interference with previous tasks. (6) \emph{Experience replay} with \emph{reservoir sampling} (ER-Res) and (7) ER with \emph{ring buffer} (ER-Ring) mitigate forgetting by directly computing gradient based on aggregated examples from new tasks and a memory buffer \citep{chaudhry2019tiny}. The ring buffer \citep{lopez2017gradient} allocates equally sized memory for each class using FIFO scheduling, while reservoir sampling \citep{vitter1985random} maintains a memory buffer by randomly drawing out already stored samples when the buffer is full. (8) \emph{Multitask} jointly learns the entire dataset in one training round, and thus is not a continual learning strategy but served as an upper bound reference for other methods.
Implementation details of the candidate methods are provided in \cref{supp:experiment}.

In our topology-based method, we regularize different aspects of the two types of neural networks employed. In the fully-connected network applied to P-MNIST and R-MNIST, we apply subgraph regularization based on two separate bipartite graphs: one consisting of neurons and weights from the two hidden layers, the other consisting of neurons and weights from the last hidden layer and the output layer. For ResNet18, we use one bipartite graph comprising neurons and weights from the pooling layer and the fully-connected output layer.
Bias neurons are not included in our regularization.
We employ topological regularization with reservoir sampling and ring buffer strategies, termed TOP-Res and TOP-Ring.
Universally, $m$ is set to 5, $p$ to 9, and $q$ to 1 in our experiment.

\paragraph{Training protocol}
We follow the training protocol of \citet{chaudhry2020continual} as follows.
The training is done universally through plain stochastic gradient descent with batch size of 10. 
All training examples in the datasets are observed only once, except for past-task examples stored in an episodic memory buffer, which can be replayed multiple times. 
We consider a tiny episodic memory with the size of one \emph{example per class} for all memory-based methods. For example, there are 100 classes in split CIFAR, and thus the memory buffer stores up to 100 past-task examples. The size of past-task batch sampled from the episodic memory is also universally set to 10, the gradient descent batch size.
Hyper-parameter tuning is performed on the first three tasks in each dataset.
A separate test set not used in the training tasks is used for performance evaluation.
Hyper-parameter tuning details are provided in \cref{supp:experiment}.

\paragraph{Performance measures}
We evaluate the algorithms based on two performance measures: average accuracy (ACC) and backward transfer (BWT) as proposed by \citet{lopez2017gradient}. ACC is the average test classification accuracy for all tasks, while BWT indicates the influence of new learning on the past knowledge, i.e., negative BWT indicates the degree of forgetting. Formally, ACC and BWT are defined as
\begin{equation}
    \text{ACC} = \frac{1}{T} \sum_{j=1}^T R_{T,j}, \quad \text{BWT} = \frac{1}{T-1} \sum_{j=1}^{T-1} R_{T,j}-R_{j,j},
\end{equation}
where $T$ is the total number of sequantial tasks, and $R_{i,j}$ is the accuracy of the model on the $j^{th}$ task after learning the $i^{th}$ task in sequence.
Higher ACC and BWT indicate better continual learning performance.
We report ACC and BWT averaged over five different task sequences for each dataset.

\begin{table}
  \caption{ACC and BWT performance for P-MNIST, R-MNIST, split CIFAR and split miniImageNet datasets. The mean and standard deviation over five different task sequences are shown.}
  \medskip
  \small
  \label{tab:results}
  \centering
  \begin{tabular}{lcrrrr}
    \toprule
        & & \multicolumn{2}{c}{P-MNIST} & \multicolumn{2}{c}{R-MNIST} \\
    \midrule
    Methods & Memory & \multicolumn{1}{c}{ACC (\%)} & \multicolumn{1}{c}{BWT (\%)} & \multicolumn{1}{c}{ACC (\%)} & \multicolumn{1}{c}{BWT (\%)} \\
    \midrule
    Finetune        & N & 34.44 $\pm$ 2.07 & -58.89 $\pm$ 2.23    & 41.43 $\pm$ 2.09 & -55.42 $\pm$ 2.09  \\
    EWC             & N & 48.05 $\pm$ 1.27 & -44.55 $\pm$ 1.46    & 39.80 $\pm$ 2.01 & -55.26 $\pm$ 2.03  \\
    RGO             & N & 73.18 $\pm$ 0.57 & -19.86 $\pm$ 0.61    & 63.34 $\pm$ 0.96 & -29.30 $\pm$ 0.95 \\
    A-GEM           & Y & 59.50 $\pm$ 1.20 & -33.63 $\pm$ 1.24    & 53.38 $\pm$ 0.90 & -43.35 $\pm$ 0.93  \\
    ORTHOG-SUB      & Y & 42.83 $\pm$ 1.22 & -9.94 $\pm$ 1.05     & 23.85 $\pm$ 0.84 & -3.45 $\pm$ 1.08 \\
    ER-Res          & Y & 66.38 $\pm$ 1.29 & -25.02 $\pm$ 1.51    & 72.54 $\pm$ 0.36 & -23.48 $\pm$ 0.39  \\
    ER-Ring         & Y & 70.10 $\pm$ 0.89 & -23.19 $\pm$ 1.01    & 70.52 $\pm$ 0.51 & -25.72 $\pm$ 0.51 \\
    TOP-Res   & Y & 68.13 $\pm$ 0.66 & -24.50 $\pm$ 0.66    & 74.33 $\pm$ 0.66 & -20.80 $\pm$ 0.63    \\
    TOP-Ring  & Y & 71.05 $\pm$ 0.80 & -22.18 $\pm$ 0.86    & 72.12 $\pm$ 0.81 & -23.85 $\pm$ 0.84    \\
    \midrule
    Multitask & -- & \multicolumn{1}{c}{90.31} & \multicolumn{1}{c}{--} & \multicolumn{1}{c}{93.43} & \multicolumn{1}{c}{--} \\
    \bottomrule
  \end{tabular}
  
  \begin{tabular}{lcrrrr}
    \toprule
        & & \multicolumn{2}{c}{Split CIFAR} & \multicolumn{2}{c}{Split miniImageNet} \\
    \midrule
    Methods & Memory & \multicolumn{1}{c}{ACC (\%)} & \multicolumn{1}{c}{BWT (\%)} & \multicolumn{1}{c}{ACC (\%)} & \multicolumn{1}{c}{BWT (\%)} \\
    \midrule
    Finetune        & N & 40.62 $\pm$ 5.09 & -23.80 $\pm$ 5.31      & 33.13 $\pm$ 2.72 & -24.95 $\pm$ 2.30 \\
    EWC             & N & 38.26 $\pm$ 3.71 & -25.30 $\pm$ 4.57      & 33.48 $\pm$ 1.79 & -19.56 $\pm$ 2.24 \\
    RGO             & N & 38.93 $\pm$ 1.03 & -18.98 $\pm$ 0.89      & 42.03 $\pm$ 1.22 & -14.19 $\pm$ 1.56 \\
    A-GEM           & Y & 43.54 $\pm$ 6.23 & -23.25 $\pm$ 5.65      & 39.52 $\pm$ 4.10 & -18.48 $\pm$ 4.39 \\
    ORTHOG-SUB      & Y & 37.93 $\pm$ 1.59 &  -5.44 $\pm$ 1.37      & 32.36 $\pm$ 1.44 &  -5.52 $\pm$ 1.07 \\
    ER-Res          & Y & 43.28 $\pm$ 1.26 & -23.08 $\pm$ 1.51      & 38.51 $\pm$ 2.40 & -13.58 $\pm$ 3.49  \\
    ER-Ring         & Y & 52.75 $\pm$ 1.18 & -14.51 $\pm$ 1.79      & 44.67 $\pm$ 1.81 & -12.23 $\pm$ 1.79 \\
    TOP-Res         & Y & 45.92 $\pm$ 1.50 & -20.10 $\pm$ 1.00      & 39.95 $\pm$ 1.91 & -14.34 $\pm$ 2.06 \\
    TOP-Ring        & Y & 54.27 $\pm$ 1.54 & -11.70 $\pm$ 1.27      & 49.08 $\pm$ 1.71 &  -8.42 $\pm$ 1.48 \\
    \midrule
    Multitask & -- & \multicolumn{1}{c}{61.08} & \multicolumn{1}{c}{--} & \multicolumn{1}{c}{57.99} & \multicolumn{1}{c}{--} \\
    \bottomrule
  \end{tabular}
\end{table}

\paragraph{Experimental results}
\cref{tab:results} shows method performance on all four datasets. \emph{Finetune} without any continual learning strategy produces lowest ACC and BWT performance, while the oracle \emph{multitask} is trained across all tasks and sets the upper bound ACC performance for all datasets. Gradient-based RGO and ORTHOG-SUB rely on over-parameterization in a neural network to reduce interference between tasks. Given the long sequence of tasks and small network architecture in our experiments, it is likely that over-parameterization is insufficient for strong performance. Although ORTHOG-SUB achieves highest BWT, it also has the lowest ACC scores, in some cases worse than \emph{Finetune}. ER-Res and ER-Ring achieve high ACC and BWT scores relative to the other baseline methods across all experiments. Our TOP-Res and TOP-Ring methods in turn demonstrate clear performance improvement over ER-Res and ER-Ring based on both ACC and BWT, suggesting that our topological continual learning strategy facilitates the consolidation of past-task knowledge beyond that provided by memory replay alone.

\begin{figure}[t]
\centering
\centerline{\includegraphics[width=1\columnwidth]{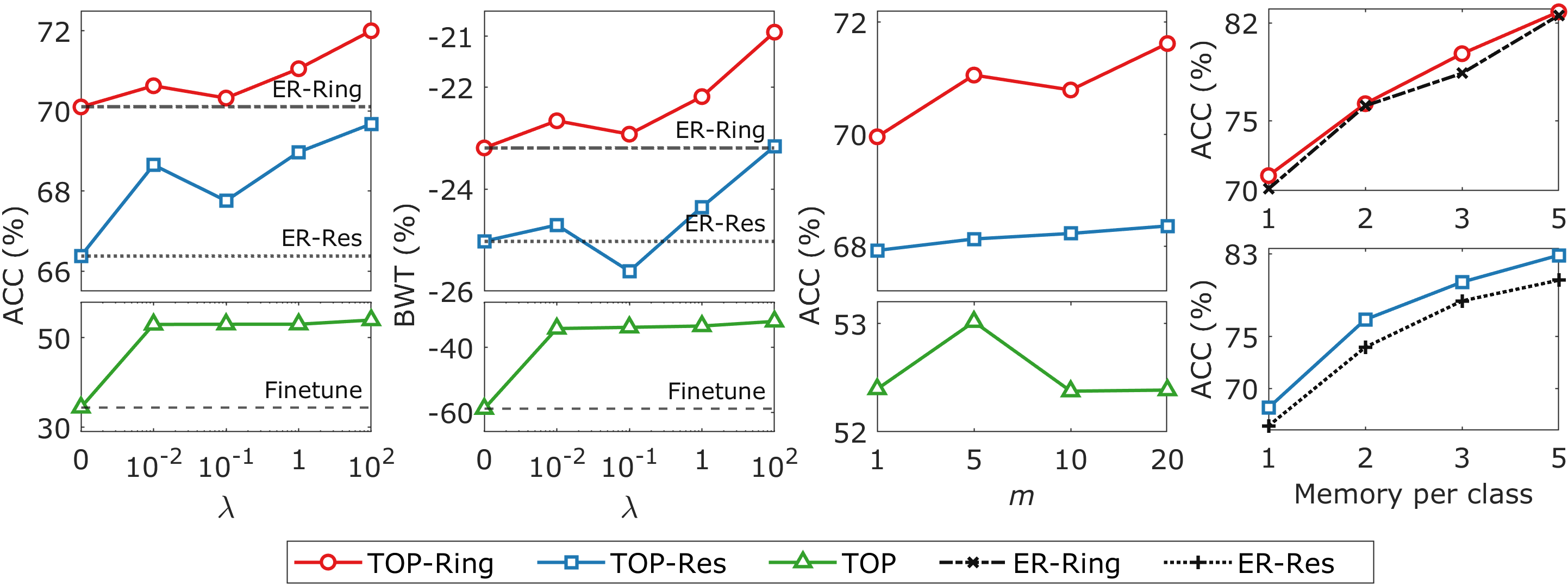}}
\caption{ACC and BWT scores as a function of the hyper-parameters of TOP-Ring, TOP-Res and TOP for P-MNIST dataset. $\lambda=0$ for TOP-Ring, TOP-Res and TOP describe ER-Ring, ER-Res and \emph{finetune}, respectively. TOP is our method without any memory buffer and uses a neural network graph constructed with the whole neural network. $\lambda$ is the weight for the topological regularization, and $m$ is the number of iterations between birth-death decomposition updates. Other hyper-parameters that are not the input to the function take the optimal values obtained by the hyper-parameter tuning.}
\label{fig:pmnist}
\end{figure}

\cref{fig:pmnist} illustrates the impact of parameters $\lambda$ and $m$ of the proposed method on the ACC and BWT measures for the P-MNIST dataset. 
The two leftmost column plots depict the increase in ACC and BWT scores provided by TOP-Ring and TOP-Res relative to ER-Ring and ER-Res baselines for all $\lambda$, the topological regularizer weight.
In addition, we demonstrate classification performance of TOP, our method without any episodic memory buffer. The results show that TOP outperforms \emph{finetune} baseline and demonstrates clear past-task knowledge retention. TOP without any memory buffer performs better than the regularization-based EWC and gradient-based ORTHOG-SUB with memory buffers, as displayed in \cref{tab:results}.
The third column plot displays ACC scores as a function of $m$, the number of iterations between birth-death decomposition updates. We observe stable upward trend as $m$ increases for TOP-Ring and TOP-Res. Less frequent birth-death decomposition update ($m=10,20$) improves performance and reduces run time. 
The last column displays ACC scores as a function of memories per class of 1, 2, 3 and 5. We observe that ACC scores increases as more past-task examples are stored in a tiny memory buffer. TOP-Ring and TOP-Res improves the performance over ER-Ring and ER-Res for all memory sizes.
Finally, we explored ACC as a function of $p/q$ of $9/1, 7/3, 5/5, 3/7$ and $1/9$, and found that the performance of both TOP-Ring and TOP-Res are not dependent on $p/q$.

\section{Discussion}
\label{sec:related}

Approaches to continual learning in neural networks can broadly be categorized into three classes. 
1) Regularization-based methods estimate importance of neural network parameters from previously learned tasks, and penalize changes to those important parameters during current tasks to mitigate forgetting \citep{kirkpatrick2017overcoming,serra2018overcoming,zenke2017continual}. Our topological regularization based method does not estimate importance for each individual parameter, but rather focuses on global network attributes. Thus, our method does not penalize individual parameters in the network, but rather penalizes cycle structure.
2) Memory-based methods attempt to overcome catastrophic forgetting by either storing examples from past tasks in a memory buffer for rehearsal \citep{chaudhry2020continual,lopez2017gradient,rebuffi2017icarl,riemer2018learning}, or synthesizing past-task data from generative models for pseudo-rehearsal \citep{shin2017continual}.
3) Expansion-based methods allocate a different set of parameters within a neural network for each task \citep{li2019learn,xu2018reinforced,yoon2018lifelong,Yoon2020Scalable}. These methods may expand network size in an attempt to eliminate forgetting by design.
In contrast, our method does not require network growth, but continually learns within a fixed network architecture.

Gradient-based methods, such as ORTHOG-SUB and RGO, can offer strong performance using over-parameterized neural networks \citep{chaudhry2020continual,liu2022continual}. However, their performance degrades substantially in a more challenging scenario of limited network width and a long sequence of learning tasks, as in our experiments. Such scenarios are especially important when computational power is limited. Large neural networks not only demand more computation to optimize, but also demand a large memory footprint to store their parameter set. Our topology-based method does not require over-parameterized neural networks to be effective and thus has limited computational burden. In particular, subgraph regularization penalizes a subset of layers towards the output side of the network and further reduces computation. Subgraph regularization gradient computation requires $n \log n$ operations and its space complexity is $n$ where $n \ll |\sW|$.
Our topology-regularized continual learning framework demonstrates the best performance in a computation constrained setting, and furthermore may provide additional insight into effective mechanisms for continual learning.

\bibliography{reference.bib}
\bibliographystyle{abbrvnat}


\newpage
\appendix
\onecolumn

\section{Birth and Death Decomposition}
\label{supp:decomp}

Persistent homology keeps track of the birth and death of topological features over filtration values $\epsilon$. If a topological feature is born at a filtration value $b_l$ and persists up to a filtration value $d_l$, then this feature is represented as a two-dimensional \emph{persistence} point $(b_l,d_l)$ in a plane. The set of all points $\{(b_l,d_l)\}_l$ is called \emph{persistence diagram} \citep{edelsbrunner2008persistent} or, equivalently, \emph{persistence barcode} \citep{ghrist2008barcodes}.
The use of the 1-skeleton simplifies the persistence barcodes to one-dimensional descriptors as follows.

The graph filtration given in (\ref{eq:graphfiltration}) begins with a graph with all parametrizing edge weights $G_{-\infty}$, sequentially removes edges at higher filtration values $\epsilon$, and arrives at an edgeless graph $G_{+\infty}$.
As $\epsilon$ increases, the number of connected components $\beta_0(G_{\epsilon})$ and cycles $\beta_1(G_{\epsilon})$ are monotonically increasing and decreasing, respectively \citep{songdechakraiwut2021topological}. Specifically, $\beta_0(G_\epsilon)$ increases from $G_{-\infty}$ consisting of a single connected component $\beta_0(G_{-\infty}) = 1$ to the node set $\beta_0(G_{\infty}) = |V|$. There are $\beta_0(G_{\infty}) - \beta_0(G_{-\infty}) = |V| - 1$ connected components that are born over the filtration. 
Once connected components are born, they will remain for all larger filtration values, so their death values are all at $+\infty$. Thus, the representation of the connected components can be simplified to a collection of sorted \emph{birth values} $\sB(G) = \{b_l \}_{l=1}^{|V|-1}$.

On the other hand, cycles are born with $G_{-\infty}$ and thus have birth values at $-\infty$. Again we can simplify the representation of the cycles as a collection of sorted \emph{death values} $\sD(G)=\{ d_l \}$.
The removal of an edge must result in either the birth of a connected component or the death of a cycle \citep{songdechakraiwut2021topological}. Thus every edge weight must also be in either $\sB(G)$ or $\sD(G)$, resulting in the decomposition of the edge weight set $\sW$ into $\sB(G)$ and $\sD(G)$.
As a result, the number of death values for cycles is equal to $|\sD(G)| = |\sW| - (|V|-1)$.
Thus, neural networks of the same architecture have a birth set $\sB$ and a death set $\sD$ of the \emph{same} cardinality as $|V| - 1$ and $|\sW| - (|V|-1)$, respectively.
This result completely resolves the problem of point mismatch in persistence barcodes for same-architecture neural networks.
Note that other filtrations \citep{carriere2020perslay,ghrist2008barcodes,hofer2020graph,otter2017roadmap,petri2013topological,zomorodian2010fast} do not necessarily share this monotonicity property. Thus, their persistence barcodes are not one-dimensional, and the number of points in the persistence barcodes may vary for same-architecture neural networks.

$\sB(G)$ comprises edge weights in the \emph{maximum spanning tree} (MST) of $G$ \citep{lee2012persistent}, and can be computed using standard methods such as Kruskal's  \citep{kruskal1956shortest} and Prim's algorithms \citep{prim1957shortest}.
Once $\sB(G)$ is identified, $\sD(G)$ is given as the remaining edge weights that are not in the MST. Thus $\sB(G)$ and $\sD(G)$ are computed very efficiently in $O(n\log n)$ operations, where $n$ is the number of edges in networks.

\section{Proofs}
\label{supp:proof}

\subsection{Closed Form Wasserstein Barycenter}

Let $\sD(G^{(i)}): d^{(i)}_{1} \leq \cdots \leq d^{(i)}_{|\sD|}$ be the death set of network $G^{(i)}$.
It follows that the $l$-th smallest death value of the barycenter $\mathcal{G}_{cycle}$ of the $N$ networks is given by the weighted mean of all the $l$-th smallest death values of such networks, i.e., $\mathcal{G}_{cycle}: \overline d_1 \leq \cdots \leq \overline d_{|\sD|},$ where
\begin{equation*}
\overline d_l = \sum_{i=1}^N \nu_i d^{(i)}_{l} \Big/ \sum_{i=1}^N \nu_i.
\end{equation*}

\begin{proof}
Recall that the \emph{Wasserstein barycenter for cycles} $\mathcal{G}_{cycle}$ is defined as the death set that minimizes the \emph{weighted} sum of the Wasserstein distances for cycles, i.e.,
\begin{align*}
\mathcal{G}_{cycle} 
&= \argmin_\mathcal{G} \sum_{i=1}^N \nu_i W^2_{cycle}(\mathcal{G}, G^{(i)}) \\
&= \argmin_\mathcal{G} \sum_{i=1}^N \nu_i \sum_{\overline d_l \in \gG} \big[\overline d_l - \phi_i^*(\overline d_l)\big]^2,
\end{align*}
where $\nu_i$ is a non-negative weight, and $\phi_i^*$ maps the $l$-th smallest death value in $\gG$ to the $l$-th smallest death value in $\sD(G^{(i)})$ for all $l$.
The sum can be expanded as
$$\sum_{i=1}^N \nu_i \sum_{\overline d_l \in \gG} \big[\overline d_l - \phi_i^*(\overline d_l)\big]^2 = \sum_{i=1}^N \nu_i \Big( [\overline d_1 - d^{(i)}_{1}]^2 + \cdots + [\overline d_{|\sD|} -d^{(i)}_{|\sD|}]^2 \Big),$$
which is quadratic. By setting its derivative equal to zero, we find the minimum at $\overline d_l = \sum_{i=1}^N \nu_i d^{(i)}_{l} \Big/ \sum_{i=1}^N \nu_i$.
\end{proof}

\subsection{Online Computation for Closed Form Wasserstein Barycenter}

Given the Wasserstein barycenter from the previous task $\gG_{cycle}^{(\tau-1)} = \{\overline d^{(\tau-1)}_{l}\}_l$.
The barycenter for the current task $\gG_{cycle}^{(\tau)} = \{\overline d^{(\tau)}_{l}\}_l$ can be computed as:
\begin{equation*}
    \overline d^{(\tau)}_{l} = \frac{p \overline d^{(\tau-1)}_{l} + q d^{(\tau)}_{l}}{p+q},
\end{equation*}
where $d^{(\tau)}_{l} \in \sD(G^{(\tau)})$ and $p,q>0$.

\begin{proof}
Let $\rho = q/(p+q)$. Then $1-\rho = p/(p+q)$. Recall $\sD(G^{(j)}) = \{d^{(j)}_{l}\}_l$ be the death set of weights after training through the entire sequence up to task $j$.

When $\tau=2$, we have the barycenter after training through the entire sequence up to the second task $\gG_{cycle}^{(2)} = \{\overline d^{(2)}_{l}\}_l,$ where
$\overline d^{(2)}_{l} = (1-\rho) d^{(1)}_{l} + \rho d^{(2)}_{l}$. That is, $\gG_{cycle}^{(2)}$ is the barycenter of $G^{(1)}$ and $G^{(2)}$ with weights associated with $G^{(1)}$ and $G^{(2)}$ as $(1-\rho)$ and $\rho$, respectively.

For $\tau>2,$ suppose $\gG_{cycle}^{(\tau-1)} = \{\overline d^{(\tau-1)}_{l}\}_l$ is the barycenter after training through the sequence up to task $\tau-1$. Then the online computation for barycenter for the next task results in $\{\overline d^{(\tau)}_{l}\}_l,$ where 
\begin{align*}
\overline d^{(\tau)}_{l} 
&= (1-\rho) \overline d^{(\tau-1)}_{l} + \rho d^{(\tau)}_{l} \\
&= (1-\rho) \sum_{i=1}^{\tau-1} \nu_i d^{(i)}_{l} \Big/ \sum_{i=1}^{\tau-1} \nu_i + \rho d^{(\tau)}_{l} \quad \nu_i>0, \forall i.
\end{align*}
It follows that the sum of weights associated with $G^{(1)},...,G^{(\tau)}$
\begin{align*}
(1-\rho)\sum_{i=1}^{\tau-1} \nu_i \Big/ \sum_{i=1}^{\tau-1} \nu_i + \rho
&= (1-\rho) \cdot 1 + \rho \\
&= 1,
\end{align*}
indicating that the $l^{th}$ smallest death value $\overline d^{(\tau)}_{l}$ is given by the weighted mean of all the $l^{th}$ smallest death values of networks $G^{(1)},...,G^{(\tau)}$. Thus, $\gG_{cycle}^{(\tau)} = \{\overline d^{(\tau)}_{l}\}_l$ is the barycenter after additional training of the $\tau^{th}$ task.
\end{proof}

\section{Experimental Details}
\label{supp:experiment}

\subsection{Implementation of Candidate Methods}

For baseline methods, we used existing implementation codes from authors' publications and publicly available repository websites.
Codes for EWC \citep{kirkpatrick2017overcoming}, A-GEM \citep{chaudhry2018efficient}, ORTHOG-SUB \citep{chaudhry2020continual}, and
ER-Res/ER-Ring \citep{chaudhry2019tiny} are available at \url{https://github.com/arslan-chaudhry/orthog_subspace} under the MIT License.
Code for RGO \citep{liu2022continual} is available at \url{https://openreview.net/forum?id=7YDLgf9_zgm}.

\subsection{Hyperparameters Tuning}

Grid search across different hypeparameter values is used to choose a set of optimal hyperparameters for all candidate methods. We consider gradient descent learning rates in $\{0.003,0.01,0.03,0.1,0.3,1.0\}$ following \citet{chaudhry2020continual,liu2022continual}. Other additional hyperparameters of EWC, A-GEM, ORTHOG-SUB, and ER-Res/ER-Ring follow \citet{chaudhry2020continual}, while those of RGO follow the authors' official implementation \citep{liu2022continual}. Specifically, below we report grids for the candidate methods with the optimal hyperparameter settings for different benchmarks given in parenthesis.

\begin{enumerate}
\small
    \item Finetune
        \begin{itemize}
            \item Learning rate: 0.003, 0.01, \textbf{0.03 (miniImageNet)}, \textbf{0.1 (P-MNIST, R-MNIST)}, \textbf{0.3 (CIFAR)}, 1
        \end{itemize}
    \item EWC
        \begin{itemize}
            \item Learning rate: 0.003, 0.01, \textbf{0.03 (CIFAR, miniImageNet)}, \textbf{0.1 (P-MNIST, R-MNIST)}, 0.3, 1
            \item Regularization: 0.01, 0.1, 1, \textbf{10 (P-MNIST, R-MNIST, CIFAR, miniImageNet)}, 100, 1000
        \end{itemize}
    \item RGO
        \begin{itemize}
            \item Learning rate: 0.003, 0.01, \textbf{0.03 (CIFAR, miniImageNet), 0.1 (P-MNIST, R-MNIST)}, 0.3, 1
        \end{itemize}
    \item A-GEM
        \begin{itemize}
            \item Learning rate: 0.003, 0.01, \textbf{0.03 (miniImageNet)}, \textbf{0.1 (P-MNIST, R-MNIST, CIFAR)}, 0.3, 1
        \end{itemize}
    \item ORTHOG-SUB
        \begin{itemize}
            \item Learning rate: 0.003, 0.01, \textbf{0.03 (R-MNIST)}, \textbf{0.1 (P-MNIST, CIFAR, miniImageNet)}, 0.3, 1
        \end{itemize}
    \item ER-Res
        \begin{itemize}
            \item Learning rate: 0.003, 0.01, \textbf{0.03 (miniImageNet)}, \textbf{0.1 (R-MNIST, CIFAR)}, \textbf{0.3 (P-MNIST)}, 1
        \end{itemize}
    \item ER-Ring
        \begin{itemize}
            \item Learning rate: 0.003, 0.01, \textbf{0.03 (miniImageNet)}, \textbf{0.1 (P-MNIST, R-MNIST, CIFAR)}, 0.3, 1
        \end{itemize}
    \item Multitask
        \begin{itemize}
            \item Learning rate: 0.003, 0.01, \textbf{0.03 (CIFAR, miniImageNet)}, \textbf{0.1 (P-MNIST, R-MNIST)}, 0.3, 1
        \end{itemize}
    \item TOP-Res
        \begin{itemize}
            \item Learning rate: 0.003, 0.01, 0.03, \textbf{0.1 (CIFAR, miniImageNet)}, \textbf{0.3 (P-MNIST, R-MNIST)}, 1
            \item Regularization: \textbf{0.01 (CIFAR, miniImageNet)}, 0.1, \textbf{1 (P-MNIST, R-MNIST)}, 10
        \end{itemize}
    \item TOP-Ring
        \begin{itemize}
            \item Learning rate: 0.003, 0.01, 0.03, \textbf{0.1 (P-MNIST, R-MNIST, CIFAR, miniImageNet)}, 0.3, 1
            \item Regularization: \textbf{0.01 (R-MNIST, miniImageNet)}, 0.1, \textbf{1 (P-MNIST, CIFAR)}, 10
        \end{itemize}
\end{enumerate}

\subsection{Computational Resources}

$1 \times$ NVIDIA GeForce GTX 1080

\end{document}

%% file: math_commands.tex
\usepackage{amsmath,amsfonts,bm}

\def\eqref#1{equation~\ref{#1}}

\def\1{\bm{1}}

\def\vx{{\bm{x}}}
\def\vy{{\bm{y}}}

\DeclareMathAlphabet{\mathsfit}{\encodingdefault}{\sfdefault}{m}{sl}
\SetMathAlphabet{\mathsfit}{bold}{\encodingdefault}{\sfdefault}{bx}{n}

\def\gG{{\mathcal{G}}}

\def\gL{{\mathcal{L}}}

\def\gX{{\mathcal{X}}}
\def\gY{{\mathcal{Y}}}

\def\sB{{\mathbb{B}}}

\def\sD{{\mathbb{D}}}

\def\sM{{\mathbb{M}}}

\def\sP{{\mathbb{P}}}

\def\sT{{\mathbb{T}}}

\def\sW{{\mathbb{W}}}

\newcommand{\E}{\mathbb{E}}

\DeclareMathOperator*{\argmin}{arg\,min}

%% file: ms.bbl
\begin{thebibliography}{51}
\providecommand{\natexlab}[1]{#1}
\providecommand{\url}[1]{\texttt{#1}}
\expandafter\ifx\csname urlstyle\endcsname\relax
  \providecommand{\doi}[1]{doi: #1}\else
  \providecommand{\doi}{doi: \begingroup \urlstyle{rm}\Url}\fi

\bibitem[Barannikov(1994)]{barannikov1994framed}
S.~Barannikov.
\newblock The framed {M}orse complex and its invariants.
\newblock \emph{Advances in Soviet Mathematics, American Mathematical Society},
  1994.

\bibitem[Bassett et~al.(2011)Bassett, Wymbs, Porter, Mucha, Carlson, and
  Grafton]{bassett2011dynamic}
D.~S. Bassett, N.~F. Wymbs, M.~A. Porter, P.~J. Mucha, J.~M. Carlson, and S.~T.
  Grafton.
\newblock Dynamic reconfiguration of human brain networks during learning.
\newblock \emph{Proceedings of the National Academy of Sciences}, 108\penalty0
  (18):\penalty0 7641--7646, 2011.
\newblock \doi{10.1073/pnas.1018985108}.

\bibitem[Bullmore and Sporns(2009)]{bullmore2009complex}
E.~Bullmore and O.~Sporns.
\newblock Complex brain networks: Graph theoretical analysis of structural and
  functional systems.
\newblock \emph{Nature Reviews Neuroscience}, 10\penalty0 (3):\penalty0
  186--198, 2009.

\bibitem[Carriere et~al.(2020)Carriere, Chazal, Ike, Lacombe, Royer, and
  Umeda]{carriere2020perslay}
M.~Carriere, F.~Chazal, Y.~Ike, T.~Lacombe, M.~Royer, and Y.~Umeda.
\newblock Perslay: {A} neural network layer for persistence diagrams and new
  graph topological signatures.
\newblock In \emph{Proceedings of the Twenty Third International Conference on
  Artificial Intelligence and Statistics}, volume 108 of \emph{Proceedings of
  Machine Learning Research}, pages 2786--2796. PMLR, 2020.

\bibitem[Chaudhry et~al.(2019{\natexlab{a}})Chaudhry, Ranzato, Rohrbach, and
  Elhoseiny]{chaudhry2018efficient}
A.~Chaudhry, M.~Ranzato, M.~Rohrbach, and M.~Elhoseiny.
\newblock Efficient lifelong learning with {A-GEM}.
\newblock In \emph{International Conference on Learning Representations},
  2019{\natexlab{a}}.

\bibitem[Chaudhry et~al.(2019{\natexlab{b}})Chaudhry, Rohrbach, Elhoseiny,
  Ajanthan, Dokania, Torr, and Ranzato]{chaudhry2019tiny}
A.~Chaudhry, M.~Rohrbach, M.~Elhoseiny, T.~Ajanthan, P.~K. Dokania, P.~H. Torr,
  and M.~Ranzato.
\newblock On tiny episodic memories in continual learning.
\newblock \emph{arXiv preprint arXiv:1902.10486}, 2019{\natexlab{b}}.

\bibitem[Chaudhry et~al.(2020)Chaudhry, Khan, Dokania, and
  Torr]{chaudhry2020continual}
A.~Chaudhry, N.~Khan, P.~Dokania, and P.~Torr.
\newblock Continual learning in low-rank orthogonal subspaces.
\newblock In \emph{Advances in Neural Information Processing Systems},
  volume~33, pages 9900--9911. Curran Associates, Inc., 2020.

\bibitem[Edelsbrunner and Harer(2008)]{edelsbrunner2008persistent}
H.~Edelsbrunner and J.~Harer.
\newblock Persistent homology--a survey.
\newblock \emph{Contemporary Mathematics}, 453:\penalty0 257--282, 2008.

\bibitem[Edelsbrunner et~al.(2000)Edelsbrunner, Letscher, and
  Zomorodian]{edelsbrunner2000topological}
H.~Edelsbrunner, D.~Letscher, and A.~Zomorodian.
\newblock Topological persistence and simplification.
\newblock In \emph{Proceedings 41st Annual Symposium on Foundations of Computer
  Science}, pages 454--463. IEEE, 2000.

\bibitem[Finc et~al.(2020)Finc, Bonna, He, Lydon-Staley, K{\"u}hn, Duch, and
  Bassett]{finc2020dynamic}
K.~Finc, K.~Bonna, X.~He, D.~M. Lydon-Staley, S.~K{\"u}hn, W.~Duch, and D.~S.
  Bassett.
\newblock Dynamic reconfiguration of functional brain networks during working
  memory training.
\newblock \emph{Nature Communications}, 11\penalty0 (1):\penalty0 1--15, 2020.

\bibitem[Ghrist(2008)]{ghrist2008barcodes}
R.~Ghrist.
\newblock Barcodes: {T}he persistent topology of data.
\newblock \emph{Bulletin of the American Mathematical Society}, 45\penalty0
  (1):\penalty0 61--75, 2008.

\bibitem[Hart and Giszter(2010)]{hart2010neural}
C.~B. Hart and S.~F. Giszter.
\newblock A neural basis for motor primitives in the spinal cord.
\newblock \emph{Journal of Neuroscience}, 30\penalty0 (4):\penalty0 1322--1336,
  2010.
\newblock \doi{10.1523/JNEUROSCI.5894-08.2010}.

\bibitem[He et~al.(2016)He, Zhang, Ren, and Sun]{he2016deep}
K.~He, X.~Zhang, S.~Ren, and J.~Sun.
\newblock Deep residual learning for image recognition.
\newblock In \emph{2016 IEEE Conference on Computer Vision and Pattern
  Recognition (CVPR)}, pages 770--778, 2016.
\newblock \doi{10.1109/CVPR.2016.90}.

\bibitem[Hofer et~al.(2020)Hofer, Graf, Rieck, Niethammer, and
  Kwitt]{hofer2020graph}
C.~Hofer, F.~Graf, B.~Rieck, M.~Niethammer, and R.~Kwitt.
\newblock Graph filtration learning.
\newblock In \emph{Proceedings of the 37th International Conference on Machine
  Learning}, volume 119 of \emph{Proceedings of Machine Learning Research},
  pages 4314--4323. PMLR, 2020.

\bibitem[Honey et~al.(2007)Honey, K{\"o}tter, Breakspear, and
  Sporns]{honey2007network}
C.~J. Honey, R.~K{\"o}tter, M.~Breakspear, and O.~Sporns.
\newblock Network structure of cerebral cortex shapes functional connectivity
  on multiple time scales.
\newblock \emph{Proceedings of the National Academy of Sciences}, 104\penalty0
  (24):\penalty0 10240--10245, 2007.

\bibitem[Kirkpatrick et~al.(2017)Kirkpatrick, Pascanu, Rabinowitz, Veness,
  Desjardins, Rusu, Milan, Quan, Ramalho, Grabska-Barwinska, Hassabis, Clopath,
  Kumaran, and Hadsell]{kirkpatrick2017overcoming}
J.~Kirkpatrick, R.~Pascanu, N.~Rabinowitz, J.~Veness, G.~Desjardins, A.~A.
  Rusu, K.~Milan, J.~Quan, T.~Ramalho, A.~Grabska-Barwinska, D.~Hassabis,
  C.~Clopath, D.~Kumaran, and R.~Hadsell.
\newblock Overcoming catastrophic forgetting in neural networks.
\newblock \emph{Proceedings of the National Academy of Sciences}, 114\penalty0
  (13):\penalty0 3521--3526, 2017.
\newblock \doi{10.1073/pnas.1611835114}.

\bibitem[Klinzing et~al.(2019)Klinzing, Niethard, and
  Born]{klinzing2019mechanisms}
J.~G. Klinzing, N.~Niethard, and J.~Born.
\newblock Mechanisms of systems memory consolidation during sleep.
\newblock \emph{Nature Neuroscience}, 22\penalty0 (10):\penalty0 1598--1610,
  2019.

\bibitem[Krizhevsky(2009)]{krizhevsky2009learning}
A.~Krizhevsky.
\newblock Learning multiple layers of features from tiny images.
\newblock Technical report, 2009.

\bibitem[Kruskal(1956)]{kruskal1956shortest}
J.~B. Kruskal.
\newblock On the shortest spanning subtree of a graph and the traveling
  salesman problem.
\newblock \emph{Proceedings of the American Mathematical society}, 7\penalty0
  (1):\penalty0 48--50, 1956.

\bibitem[Kwon and Cho(2007)]{kwon2007analysis}
Y.-K. Kwon and K.-H. Cho.
\newblock Analysis of feedback loops and robustness in network evolution based
  on boolean models.
\newblock \emph{BMC Bioinformatics}, 8\penalty0 (1):\penalty0 1--9, 2007.

\bibitem[Lecun et~al.(1998)Lecun, Bottou, Bengio, and
  Haffner]{lecun1998gradient}
Y.~Lecun, L.~Bottou, Y.~Bengio, and P.~Haffner.
\newblock Gradient-based learning applied to document recognition.
\newblock \emph{Proceedings of the IEEE}, 86\penalty0 (11):\penalty0
  2278--2324, 1998.
\newblock \doi{10.1109/5.726791}.

\bibitem[Lee et~al.(2012)Lee, Kang, Chung, Kim, and Lee]{lee2012persistent}
H.~Lee, H.~Kang, M.~K. Chung, B.-N. Kim, and D.~S. Lee.
\newblock Persistent brain network homology from the perspective of dendrogram.
\newblock \emph{IEEE Transactions on Medical Imaging}, 31\penalty0
  (12):\penalty0 2267--2277, 2012.
\newblock \doi{10.1109/TMI.2012.2219590}.

\bibitem[Li et~al.(2019)Li, Zhou, Wu, Socher, and Xiong]{li2019learn}
X.~Li, Y.~Zhou, T.~Wu, R.~Socher, and C.~Xiong.
\newblock Learn to grow: {A} continual structure learning framework for
  overcoming catastrophic forgetting.
\newblock In \emph{International Conference on Machine Learning}, pages
  3925--3934. PMLR, 2019.

\bibitem[Liu and Liu(2022)]{liu2022continual}
H.~Liu and H.~Liu.
\newblock Continual learning with recursive gradient optimization.
\newblock In \emph{International Conference on Learning Representations}, 2022.

\bibitem[Lopez-Paz and Ranzato(2017)]{lopez2017gradient}
D.~Lopez-Paz and M.~A. Ranzato.
\newblock Gradient episodic memory for continual learning.
\newblock In \emph{Advances in Neural Information Processing Systems},
  volume~30. Curran Associates, Inc., 2017.

\bibitem[McCloskey and Cohen(1989)]{mccloskey1989catastrophic}
M.~McCloskey and N.~J. Cohen.
\newblock Catastrophic interference in connectionist networks: {T}he sequential
  learning problem.
\newblock volume~24 of \emph{Psychology of Learning and Motivation}, pages
  109--165. Academic Press, 1989.
\newblock \doi{10.1016/S0079-7421(08)60536-8}.

\bibitem[McGaugh(2000)]{mcgaugh2000memory}
J.~L. McGaugh.
\newblock Memory--a century of consolidation.
\newblock \emph{Science}, 287\penalty0 (5451):\penalty0 248--251, 2000.

\bibitem[Munkres(1996)]{munkres1996elements}
J.~R. Munkres.
\newblock \emph{Elements Of Algebraic Topology}.
\newblock Avalon Publishing, 1996.
\newblock ISBN 9780201627282.

\bibitem[Otter et~al.(2017)Otter, Porter, Tillmann, Grindrod, and
  Harrington]{otter2017roadmap}
N.~Otter, M.~A. Porter, U.~Tillmann, P.~Grindrod, and H.~A. Harrington.
\newblock A roadmap for the computation of persistent homology.
\newblock \emph{EPJ Data Science}, 6, 2017.
\newblock \doi{10.1140/epjds/s13688-017-0109-5}.

\bibitem[Ozbudak et~al.(2005)Ozbudak, Becskei, and
  Van~Oudenaarden]{ozbudak2005system}
E.~M. Ozbudak, A.~Becskei, and A.~Van~Oudenaarden.
\newblock A system of counteracting feedback loops regulates {Cdc42p} activity
  during spontaneous cell polarization.
\newblock \emph{Developmental Cell}, 9\penalty0 (4):\penalty0 565--571, 2005.

\bibitem[Petri et~al.(2013)Petri, Scolamiero, Donato, and
  Vaccarino]{petri2013topological}
G.~Petri, M.~Scolamiero, I.~Donato, and F.~Vaccarino.
\newblock Topological strata of weighted complex networks.
\newblock \emph{PLOS ONE}, 8\penalty0 (6):\penalty0 1--8, 2013.

\bibitem[Prim(1957)]{prim1957shortest}
R.~C. Prim.
\newblock Shortest connection networks and some generalizations.
\newblock \emph{The Bell System Technical Journal}, 36\penalty0 (6):\penalty0
  1389--1401, 1957.

\bibitem[Rebuffi et~al.(2017)Rebuffi, Kolesnikov, Sperl, and
  Lampert]{rebuffi2017icarl}
S.-A. Rebuffi, A.~Kolesnikov, G.~Sperl, and C.~H. Lampert.
\newblock {iCaRL}: {I}ncremental classifier and representation learning.
\newblock In \emph{Proceedings of the IEEE conference on Computer Vision and
  Pattern Recognition}, pages 2001--2010, 2017.

\bibitem[Riemer et~al.(2019)Riemer, Cases, Ajemian, Liu, Rish, Tu, , and
  Tesauro]{riemer2018learning}
M.~Riemer, I.~Cases, R.~Ajemian, M.~Liu, I.~Rish, Y.~Tu, , and G.~Tesauro.
\newblock Learning to learn without forgetting by maximizing transfer and
  minimizing interference.
\newblock In \emph{International Conference on Learning Representations}, 2019.

\bibitem[Russakovsky et~al.(2015)Russakovsky, Deng, Su, Krause, Satheesh, Ma,
  Huang, Karpathy, Khosla, Bernstein, Berg, and
  Fei-Fei]{russakovsky2015imagenet}
O.~Russakovsky, J.~Deng, H.~Su, J.~Krause, S.~Satheesh, S.~Ma, Z.~Huang,
  A.~Karpathy, A.~Khosla, M.~Bernstein, A.~C. Berg, and L.~Fei-Fei.
\newblock {ImageNet} large scale visual recognition challenge.
\newblock \emph{International Journal of Computer Vision}, 115\penalty0
  (3):\penalty0 211--252, 2015.

\bibitem[Serra et~al.(2018)Serra, Suris, Miron, and
  Karatzoglou]{serra2018overcoming}
J.~Serra, D.~Suris, M.~Miron, and A.~Karatzoglou.
\newblock Overcoming catastrophic forgetting with hard attention to the task.
\newblock In \emph{International Conference on Machine Learning}, pages
  4548--4557. PMLR, 2018.

\bibitem[Shin et~al.(2017)Shin, Lee, Kim, and Kim]{shin2017continual}
H.~Shin, J.~K. Lee, J.~Kim, and J.~Kim.
\newblock Continual learning with deep generative replay.
\newblock In \emph{Advances in Neural Information Processing Systems},
  volume~30, 2017.

\bibitem[Simon(1962)]{simon1962architecture}
H.~A. Simon.
\newblock The architecture of complexity.
\newblock \emph{Proceedings of the American Philosophical Society}, 106, 1962.

\bibitem[Songdechakraiwut et~al.(2021)Songdechakraiwut, Shen, and
  Chung]{songdechakraiwut2021topological}
T.~Songdechakraiwut, L.~Shen, and M.~Chung.
\newblock Topological learning and its application to multimodal brain network
  integration.
\newblock In \emph{Medical Image Computing and Computer Assisted Intervention
  -- MICCAI 2021}, pages 166--176, Cham, 2021. Springer International
  Publishing.
\newblock ISBN 978-3-030-87196-3.

\bibitem[Songdechakraiwut et~al.(2022)Songdechakraiwut, Krause, Banks, Nourski,
  and Veen]{songdechakraiwut2022fast}
T.~Songdechakraiwut, B.~M. Krause, M.~I. Banks, K.~V. Nourski, and B.~D.~V.
  Veen.
\newblock Fast topological clustering with {Wasserstein} distance.
\newblock In \emph{International Conference on Learning Representations}, 2022.
\newblock URL \url{https://openreview.net/forum?id=0kPL3xO4R5}.

\bibitem[Vapnik(1991)]{vapnik1991principles}
V.~Vapnik.
\newblock Principles of risk minimization for learning theory.
\newblock In \emph{Advances in Neural Information Processing Systems},
  volume~4. Morgan-Kaufmann, 1991.

\bibitem[Venkatesh et~al.(2004)Venkatesh, Bhartiya, and
  Ruhela]{venkatesh2004multiple}
K.~Venkatesh, S.~Bhartiya, and A.~Ruhela.
\newblock Multiple feedback loops are key to a robust dynamic performance of
  tryptophan regulation in {E}scherichia coli.
\newblock \emph{FEBS Letters}, 563\penalty0 (1-3):\penalty0 234--240, 2004.

\bibitem[Vinyals et~al.(2016)Vinyals, Blundell, Lillicrap, kavukcuoglu, and
  Wierstra]{vinyals2016matching}
O.~Vinyals, C.~Blundell, T.~Lillicrap, k.~kavukcuoglu, and D.~Wierstra.
\newblock Matching networks for one shot learning.
\newblock In \emph{Advances in Neural Information Processing Systems},
  volume~29. Curran Associates, Inc., 2016.

\bibitem[Vitter(1985)]{vitter1985random}
J.~S. Vitter.
\newblock Random sampling with a reservoir.
\newblock \emph{ACM Trans. Math. Softw.}, 11\penalty0 (1):\penalty0 37–57,
  1985.
\newblock \doi{10.1145/3147.3165}.

\bibitem[Wasserman(2018)]{wasserman2018topological}
L.~Wasserman.
\newblock Topological data analysis.
\newblock \emph{Annual Review of Statistics and Its Application}, 5:\penalty0
  501--532, 2018.

\bibitem[Weiner et~al.(2002)Weiner, Neilsen, Prestwich, Kirschner, Cantley, and
  Bourne]{weiner2002ptdinsp}
O.~D. Weiner, P.~O. Neilsen, G.~D. Prestwich, M.~W. Kirschner, L.~C. Cantley,
  and H.~R. Bourne.
\newblock A {PtdInsP} 3-and {Rho} {GTPase}-mediated positive feedback loop
  regulates neutrophil polarity.
\newblock \emph{Nature Cell Biology}, 4\penalty0 (7):\penalty0 509--513, 2002.

\bibitem[Xu and Zhu(2018)]{xu2018reinforced}
J.~Xu and Z.~Zhu.
\newblock Reinforced continual learning.
\newblock In \emph{Advances in Neural Information Processing Systems},
  volume~31, 2018.

\bibitem[Yoon et~al.(2018)Yoon, Yang, Lee, and Hwang]{yoon2018lifelong}
J.~Yoon, E.~Yang, J.~Lee, and S.~J. Hwang.
\newblock Lifelong learning with dynamically expandable networks.
\newblock In \emph{International Conference on Learning Representations}, 2018.

\bibitem[Yoon et~al.(2020)Yoon, Kim, Yang, and Hwang]{Yoon2020Scalable}
J.~Yoon, S.~Kim, E.~Yang, and S.~J. Hwang.
\newblock Scalable and order-robust continual learning with additive parameter
  decomposition.
\newblock In \emph{International Conference on Learning Representations}, 2020.

\bibitem[Zenke et~al.(2017)Zenke, Poole, and Ganguli]{zenke2017continual}
F.~Zenke, B.~Poole, and S.~Ganguli.
\newblock Continual learning through synaptic intelligence.
\newblock In \emph{International Conference on Machine Learning}, pages
  3987--3995. PMLR, 2017.

\bibitem[Zomorodian(2010)]{zomorodian2010fast}
A.~Zomorodian.
\newblock Fast construction of the {Vietoris-Rips} complex.
\newblock \emph{Computers \& Graphics}, 34\penalty0 (3):\penalty0 263--271,
  2010.

\end{thebibliography}
